\documentclass[11pt,letterpaper]{mystyle}

\usepackage{fancyhdr}
\usepackage{float}
\usepackage[utf8]{inputenc}
\usepackage[T1]{fontenc}
\usepackage[numbers]{natbib}
\usepackage{hyperref}
\usepackage{mathtools}
\usepackage{bm}
\usepackage{wrapfig}
\usepackage{bxcoloremoji}
\usepackage{CJKutf8}
\usepackage{placeins}
\usepackage{algorithm}
\usepackage{algpseudocode}
\usepackage{array}

\usepackage{amsmath,amsfonts,bm}

\def\eqref#1{equation~\ref{#1}}

\def\1{\bm{1}}

\DeclareMathAlphabet{\mathsfit}{\encodingdefault}{\sfdefault}{m}{sl}
\SetMathAlphabet{\mathsfit}{bold}{\encodingdefault}{\sfdefault}{bx}{n}

\newcommand{\github}{\raisebox{-1.5pt}{\includegraphics[height=1.05em]{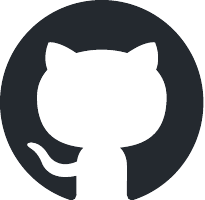}}}
\DeclareRobustCommand{\comment}[1]{%
  \begingroup\normalfont
  \begin{CJK*}{UTF8}{gbsn}\textcolor{red}{[#1]}\end{CJK*}%
  \endgroup
}
\definecolor{dmpddelta}{HTML}{4F7A57}
\definecolor{dmpdnegative}{HTML}{A45D52}
\definecolor{darkblue}{rgb}{0,0.40,0.75}
\newcolumntype{Y}{>{\raggedright\arraybackslash}X}
\graphicspath{{figures/}}
\hypersetup{
  colorlinks=true,citecolor=darkblue,linkcolor=darkblue,urlcolor=darkblue,
  pdftitle={ATI-OPD: Adaptive Teacher Intervention for Stable Long-Sequence On-Policy Distillation},
  pdfauthor={Jiannan Guan, Youxin Jiang, Yutai Hou, Qiguang Chen, Yirong Zeng, Haitao Wang, Dengyun Peng, Zheng Yan, Jinhao Liu, Tao Ye, Dandan Tu, Wanxiang Che}
}

\fancypagestyle{headstyle}{%
  \fancyhead[L]{\includegraphics[width=80pt]{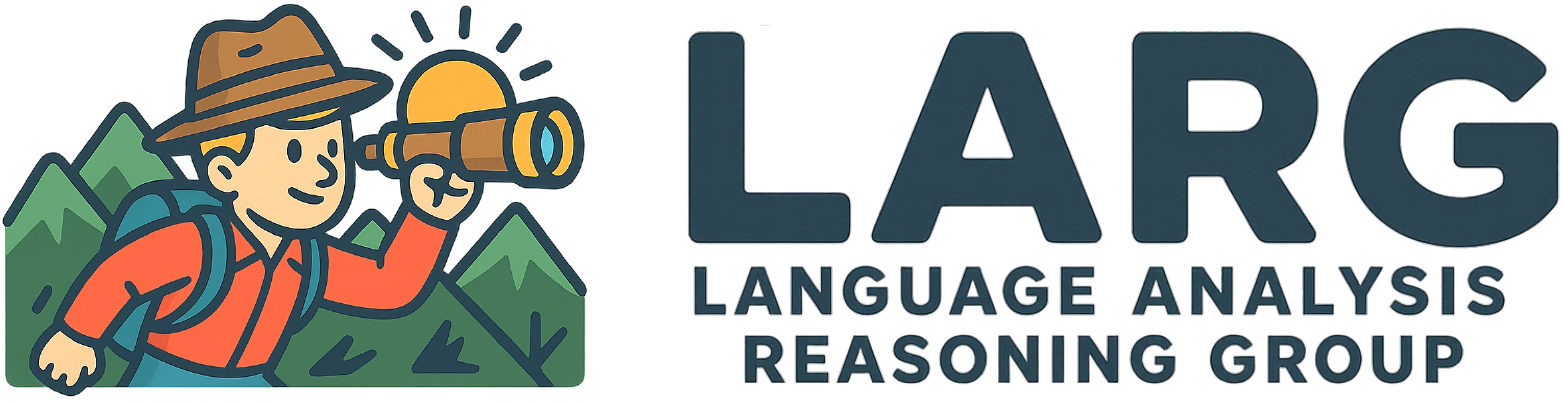}}
  \fancyhead[C]{}
  \fancyhead[R]{}
}

\title{ATI-OPD: Adaptive Teacher Intervention for \\ Stable Long-Sequence On-Policy Distillation}

\author{
  Jiannan Guan$^{1*}$ \quad
  Youxin Jiang$^{1*}$ \quad
  Yutai Hou$^{2}$ \quad
  Qiguang Chen$^{1}$ \quad
  Yirong Zeng$^{1}$ \quad
  Haitao Wang$^{2}$ \quad
  Dengyun Peng$^{1}$ \quad
  Zheng Yan$^{1}$ \quad
  Jinhao Liu$^{1}$ \quad
  Tao Ye$^{2}$ \quad
  Dandan Tu$^{2}$ \quad
  Wanxiang Che$^{1, \protect\coloremojicode{2709}}$ \vspace{-5pt}\\
  \normalfont{
  $^1$ Research Center for Social Computing and Interactive Robotics, Harbin Institute of Technology,\vspace{-5pt}\\
  $^2$ Huawei Technologies Co., Ltd. \\
  }
}

\begin{document}

\begin{abstract}
  \vspace{5mm}
  \textbf{\large Abstract:}
  \vspace{2mm}

On-policy distillation (OPD) trains a student on its own responses using token-level feedback from a teacher.
As responses grow longer, the student's prefix can depart from the teacher's own reasoning, weakening its feedback on later tokens.
Updates based on this feedback can reinforce the same deviations in later responses and destabilize training.
Prior approaches such as length control and loss reweighting limit learning from problematic responses but leave their prefixes unchanged. Fixed cutoffs can also discard later reasoning.
We propose \textbf{Adaptive Teacher Intervention for On-Policy Distillation (ATI-OPD)} to improve subsequent teacher guidance by revising the student's reasoning prefix.
ATI-OPD replaces the earliest high-disagreement segment with a short teacher segment, then lets the student resume its own reasoning from the revised text.
To handle renewed disagreement, it excludes persistently divergent suffixes from the distillation loss while retaining useful student continuation.
Training diagnostics show less teacher-student disagreement, fewer gradient spikes, and fewer extremely long responses.
Across five mathematical reasoning benchmarks and two teacher-student pairs, ATI-OPD improves average student-only accuracy over standard OPD.
  \vspace{5mm}

  $^{*}$ \textit{Equal Contribution}

  $^{\coloremojicode{2709}}$ \textit{Corresponding Author}

  \vspace{5mm}

  \github{} \textbf{Code Repository}: \href{https://github.com/jubgjf/ATI-OPD}{https://github.com/jubgjf/ATI-OPD}

  \coloremojicode{1F4E7} \textbf{Contact}: \href{mailto:jnguan@ir.hit.edu.cn}{jnguan@ir.hit.edu.cn}, \href{mailto:yxjiang@ir.hit.edu.cn}{yxjiang@ir.hit.edu.cn}, \href{mailto:car@ir.hit.edu.cn}{car@ir.hit.edu.cn}

\end{abstract}
\maketitle

\vspace{3mm}
\pagestyle{headstyle}
\thispagestyle{empty}

\begin{CJK}{UTF8}{gbsn}

\begin{figure}[t]
    \centering
    \includegraphics[width=0.8\linewidth]{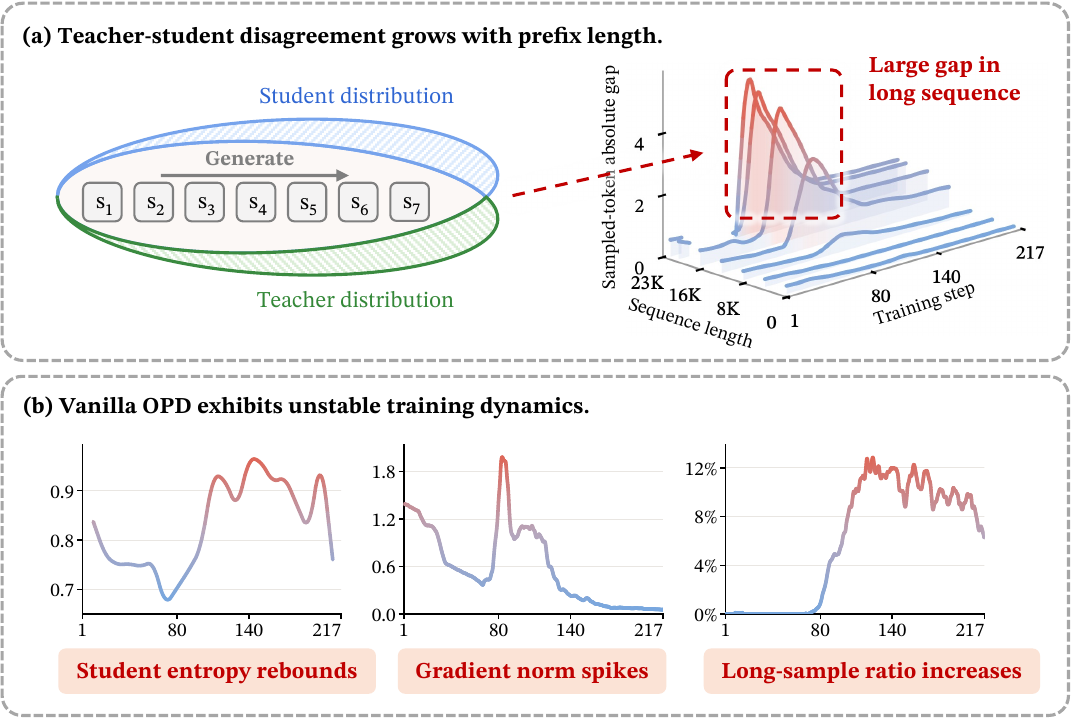}
    \caption{
        Vanilla OPD on JustRL-DeepSeek exhibits prediction disagreement and training instability.
        (a) Gap measures the absolute difference between teacher and student log-probabilities for the same sampled token and prefix.
        (b) Across training steps, entropy rebounds, gradient norms spike, and more responses reach the max token limit.
    }
    \label{fig:motivation}
\end{figure}

\section{Introduction}
\label{sec:introduction}

On-policy distillation (OPD) trains a student on responses it generates itself, using a teacher's token-level predictions as supervision~\citep{agarwal2024onpolicy,gu2024minillm}.
Learning from its own prefixes exposes the student to the decisions it must make at inference.
The teacher must therefore provide useful guidance from the prefixes that the student actually generates.

Generating long sequences can weaken this guidance and make training unstable.
As the prefix grows, it can depart from the reasoning that the teacher would generate on its own.
Even a strong teacher can become less effective when continuing student-written prefixes~\citep{zhu2026manyfaces}.
OPD uses these token-level predictions to update the student, so their quality affects which continuations it learns to generate.
Existing works have found that repetitive continuations can receive favorable local updates.
As the student samples them more often, they contribute more heavily to subsequent updates and reinforce length inflation~\citep{luo2026demystifyingopd}.
In long-sequence OPD, declining teacher-student agreement has also been observed alongside greater prediction uncertainty and gradient spikes~\citep{li2026rethinkingopd}.
We measure disagreement with the \emph{sampled-token absolute gap}: the absolute difference between teacher and student log-probabilities for a sampled token under the same prefix.
In vanilla OPD, large gaps appear at later response positions.
Over training steps, entropy rebounds, gradient norms spike, and more responses reach the token limit, as shown in Figure~\ref{fig:motivation}.

To reduce the impact of unreliable feedback, prior methods control response length or reweight token losses~\citep{ziheng2026less,zhang2026fullrollouts,xie2026positionbias}.
These approaches limit learning from problematic continuations but leave the generated prefix unchanged.
Fixed cutoffs can also discard useful later reasoning.
Teacher participation can instead change the course of generation through token replacement or local takeover~\citep{xu2025speculative,koo2025switch,xu2026pass}.
However, improving the prefix does not ensure that the student's later decisions remain useful for learning: disagreement can return after generation is handed back to the student.
This calls for local guidance that changes the reasoning trajectory, together with supervision selection based on the regenerated continuation.

To address these challenges, we propose \textbf{Adaptive Teacher Intervention for On-Policy Distillation (ATI-OPD)} to revise the student's reasoning prefix and select supervision within the resulting continuation.
Its core module, \textbf{Adaptive Teacher Intervention}, locates the earliest high-disagreement segment in the student response.
The teacher takes over from the preceding prefix and replaces that segment, allowing the student to regenerate what follows.
Targeting the earliest high-disagreement segment changes the prefix before that segment influences later decisions.
A short teacher segment guides further reasoning while keeping the student's own decisions in the training trajectory.

The student can diverge again as it continues beyond the local teacher rewrite.
\textbf{Adaptive Supervision Selection} therefore monitors successive windows of the new continuation and excludes persistently divergent suffixes from the distillation loss.
This retains useful student supervision beyond the teacher segment while limiting updates on later disagreement.

Our contributions are as follows:

\begin{enumerate}[leftmargin=16pt, itemsep=3pt, topsep=3pt]
    \item We introduce Adaptive Teacher Intervention to address weakened guidance on student-generated prefixes. Replacing a high-disagreement segment with a short teacher segment improves alignment in the student's continuation.
    \item We introduce Adaptive Supervision Selection to handle renewed disagreement after intervention. Selective supervision outperforms both learning from the full suffix, even with more interventions, and stopping supervision at the teacher segment.
    \item ATI-OPD reduces the area under the absolute-gap curve (AUC) by $47.8\%$ and suppresses gradient spikes and extremely long responses during training. More stable training is accompanied by higher average student-only accuracy across five benchmarks on both model pairs.
\end{enumerate}

\section{Related Work}
\label{sec:related_work}

\paragraph{OPD objectives and feedback stability.}
OPD methods improve feedback by changing divergence objectives~\citep{agarwal2024onpolicy,ko2024distillm}, forming a weighted product of teacher and student distributions as the supervision target~\citep{jang2026stableopd}, or adding supervision at high-entropy teacher positions~\citep{jin2026entropyaware}.
Stable-OPD addresses a feedback loop between repetitive continuations and favorable local updates through off-policy supervision and reference-policy regularization~\citep{luo2026demystifyingopd}.
ATI-OPD changes trajectories and supervision ranges while retaining the OPD update.

\paragraph{Teacher participation in trajectory generation.}
Teacher participation ranges from mixing sampling distributions~\citep{gu2024minillm} and selecting or replacing individual tokens~\citep{xu2025speculative,koo2025switch} to taking over the remaining suffix~\citep{peng2025adaswitch}.
Teacher rewriting also appears in MOTAB, which joins corrective teacher suffixes to erroneous student prefixes~\citep{wang2026backtracking}, and TRD, which asks the teacher to revise completed responses~\citep{jiang2026trajectory}.
Relay-OPD triggers takeovers when the teacher prefers a reflection token absent from the student's leading predictions and ends the rollout after its final teacher segment~\citep{xu2026pass}.
ATI-OPD instead triggers on segment-averaged divergence and lets the student complete the response after a bounded replacement.
Its supervision cutoff depends on persistent divergence in that continuation, rather than on the number of teacher takeovers.

\paragraph{Length control and selective supervision.}
Fixed-prefix and progressive-length methods restrict rollout lengths without using disagreement in the current continuation~\citep{zhang2026prefixes,ziheng2026less,zhang2026fullrollouts}.
Prune-OPD reduces rewards after low-overlap events and adjusts the next training step's rollout budget using batch-level reliable lengths~\citep{yang2026prune}, whereas KAT stops and rolls back rollouts after persistent low-KL agreement, calibrating its threshold from recent rollout minima~\citep{xin2026escaping}.
IW-OPD reweights compatible prefixes while retaining dense supervision~\citep{xie2026positionbias}.
AC-OPD retains the original student trajectory and adds teacher-generated branches, selecting each branch's supervision length using uncertainty and accumulated divergence~\citep{Zhou2026TeacherST}.
ATI-OPD instead replaces a student segment and selects which tokens in the regenerated student suffix receive OPD supervision, with its threshold calibrated from correct original responses.
This selection takes place after full generation and changes only the loss mask.

\section{Preliminaries}
\label{sec:preliminaries}

\subsection{On-policy distillation}
\label{sec:opd}

Let $\pi_\theta$ and $\pi_T$ denote the student and fixed teacher.
For question $x$, OPD samples $y=(y_1,\ldots,y_N)$ from a student snapshot $\pi_{\theta_{\mathrm{old}}}$ held fixed during batch generation.
With $h_t=(x,y_{<t})$ denoting the question and preceding text, the reverse-KL objective is
\begingroup
\postdisplaypenalty=10000 
\begin{equation}
\mathcal L_{\mathrm{OPD}}(\theta)
=
\mathbb E_{x,y}
\left[
\frac{1}{N}\sum_{t=1}^{N}
D_{\mathrm{KL}}\!\left(
\pi_\theta(\cdot\mid h_t)\Vert\pi_T(\cdot\mid h_t)
\right)
\right].
\label{eq:opd}
\end{equation}
\endgroup

Here, $N$ is the response length, and the expectation samples questions from the training distribution and responses from the rollout student.
In practice, we query exact teacher log-probabilities for the rollout student's top-16 candidates and use a clipped policy surrogate~\citep{li2026rethinkingopd}.
See Appendix~\ref{app:exact_opd_surrogate} for the implementation details.

\subsection{Sampled-token divergence}
\label{sec:state_drift}

We monitor teacher-student disagreement using a sampled-token estimator~\citep{schulman2020kl}:
\begingroup
\postdisplaypenalty=10000
\begin{equation}
\delta_t=\log\pi_T(y_t\mid h_t)
-\log\pi_{\theta_{\mathrm{old}}}(y_t\mid h_t),
\qquad
g_t=\exp(\delta_t)-\delta_t-1 .
\label{eq:k3}
\end{equation}
\endgroup
Here, $\delta_t$ is the log-probability difference and $g_t$ is the nonnegative sampled-token estimator, commonly called K3.
With full support and no clipping,
$\mathbb E_{y_t\sim\pi_{\theta_{\mathrm{old}}}}[g_t]
=D_{\mathrm{KL}}(\pi_{\theta_{\mathrm{old}}}\Vert\pi_T)$.
We clip extreme values for numerical stability.
Monitoring and local K3 diagnostics use the resulting score $\bar g_t$. See Appendix~\ref{app:trajectory_implementation} for details.

\section{Adaptive Teacher Intervention for On-Policy Distillation}
\label{sec:dmpd}

\begin{figure}[t]
    \centering
    \includegraphics[width=0.9\linewidth]{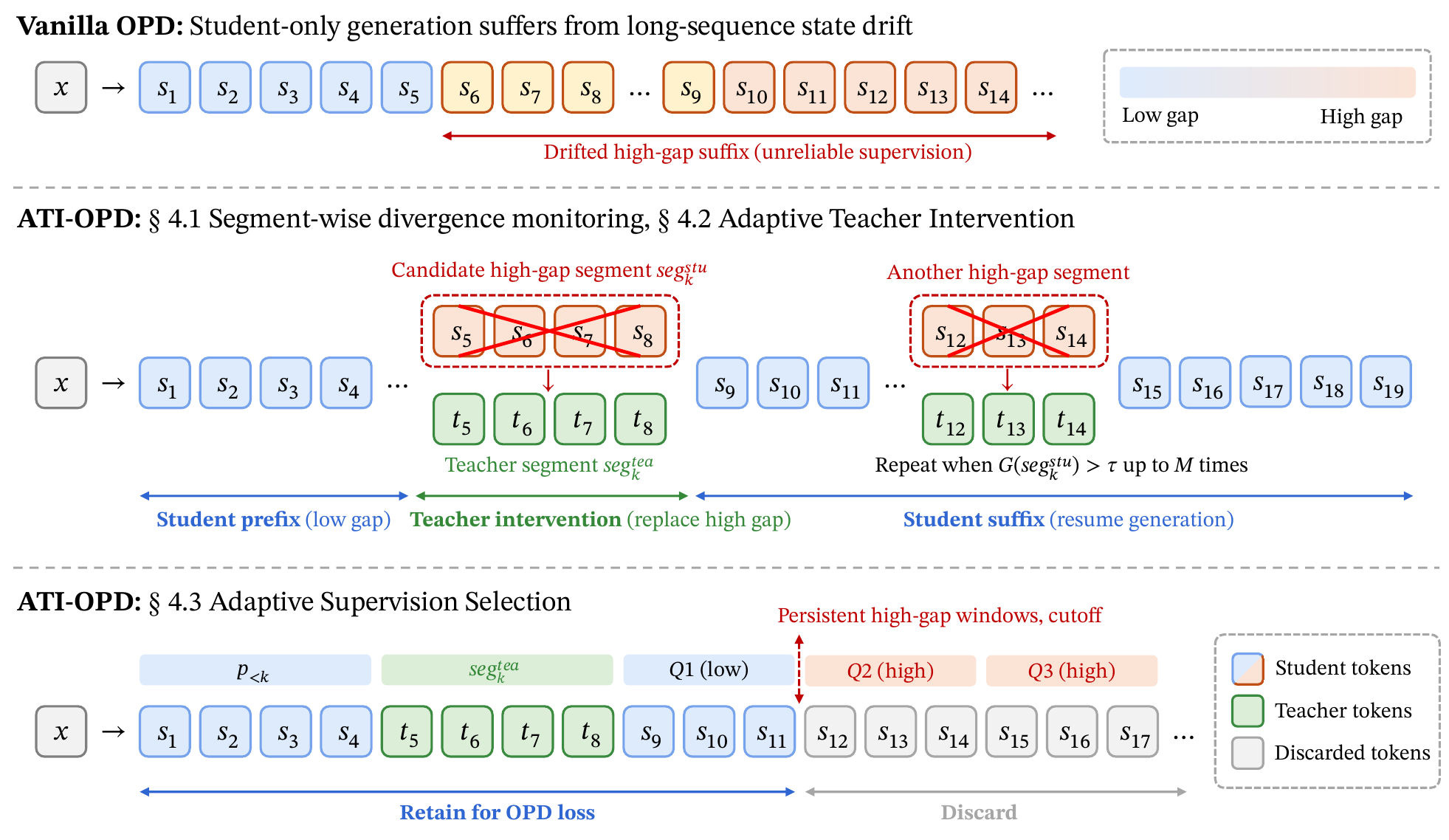}
    \caption{
        ATI-OPD replaces a high-divergence student segment with a teacher segment and lets the student generate a new continuation. Persistent divergence after the handoff determines a supervision cutoff. Later tokens remain in the generated response but are excluded from distillation.
        }
    \label{fig:dmpd_overview}
\end{figure}

ATI-OPD links where teacher guidance changes student generation with how far the new continuation contributes to learning.
It first locates an intervention segment, rewrites it with the teacher, and then selects subsequent supervision positions, as shown in Figure~\ref{fig:dmpd_overview}.

\subsection{Segment-wise Divergence Monitoring}
\label{sec:dmpd_monitor}

We first generate a complete student response $y^{(0)}$ and score it with the teacher.
To locate where their predictions diverge, we divide the response into segments of at most $L_S$ tokens and compute
\begingroup
\postdisplaypenalty=10000
\begin{equation}
    G(\mathrm{seg}_k^{\mathrm{stu}})
    =\frac{\sum_{t\in\mathrm{seg}_k^{\mathrm{stu}}}m_t \bar g_t}
    {\sum_{t\in\mathrm{seg}_k^{\mathrm{stu}}}m_t}.
    \label{eq:segment_gap}
\end{equation}
\endgroup
Here, $\mathrm{seg}_k^{\mathrm{stu}}$ is student segment $k$, and $m_t$ marks valid student tokens, excluding padding and teacher tokens.
We scan in response order and intervene at the first segment whose mean exceeds threshold $\tau$, if budget remains.
The main setting uses $\tau=0.15$.
Appendix~\ref{app:trigger_threshold_sensitivity} examines its sensitivity.

\subsection{Adaptive Teacher Intervention}
\label{sec:dmpd_repair}

To give the student a revised starting point, we retain the prefix $p_{<k}$ before the triggering segment, discard that segment and the original suffix, and sample a bounded teacher replacement:
\begingroup
\postdisplaypenalty=10000
\begin{equation}
    \mathrm{seg}_k^{\mathrm{tea}}\sim\pi_T(\cdot\mid x,p_{<k}),
    \qquad |\mathrm{seg}_k^{\mathrm{tea}}|\le L_T.
    \label{eq:state_repair}
\end{equation}
\endgroup
Here, $L_T$ limits the teacher segment length, and $|\cdot|$ denotes sequence length.
Unless the response has ended, the student continues from $(x,p_{<k},\mathrm{seg}_k^{\mathrm{tea}})$ until EOS or the total response budget $N_{\max}$.
The retained prefix, teacher segment, and complete new suffix form a mixed-policy trajectory $\widetilde y$.

\subsection{Adaptive Supervision Selection}
\label{sec:dmpd_supervision}

A local intervention does not ensure useful guidance throughout the remaining response.
We therefore score the complete new student continuation in windows of length $L_F$ and select its supervision range using
\begingroup
\postdisplaypenalty=10000
\begin{equation}
    Q_j=\frac{1}{|W_j|}\sum_{t\in W_j}\bar g_t,
    \qquad
    \eta_s=\operatorname{Quantile}_{0.90}(\mathcal H_s).
    \label{eq:eta_s}
\end{equation}
\endgroup
Here, $W_j$ contains the valid student tokens in window $j$, and $|W_j|$ is their count.
At training step $s$, the reference pool $\mathcal H_s$ contains window means from earlier original responses with correct final answers.
These windows provide an empirical scale of disagreement during successful student generation, allowing threshold $\eta_s$ to follow changes in the student.
We use $\eta_s=\tau$ until enough reference windows are available. See Appendix~\ref{app:hyperparameters} for details.

Two consecutive valid windows above $\eta_s$ trigger the supervision cutoff:
\begingroup
\postdisplaypenalty=10000 
\begin{equation}
    j^\star=\min\{j:Q_j>\eta_s,\ Q_{j+1}>\eta_s\},
    \qquad
    c=
    \begin{cases}
        b_{j^\star}-1, & \text{if such a pair exists},\\
        N, & \text{otherwise}.
    \end{cases}
    \label{eq:cutoff_condition}
\end{equation}
\endgroup
Here, $j^\star$ indexes the first violating window pair, $b_j$ is the first position of window $j$, and $N$ is the complete mixed response length.
The cutoff position $c$ is the last token retained for supervision.
A loss mask excludes both violating windows and all later tokens, without changing the generated text.
The retained student prefix, teacher segment, and selected continuation receive the same OPD update as in Section~\ref{sec:opd}.
Appendix~\ref{app:trajectory_implementation} gives the detailed loss mask implementation and ATI-OPD full algorithm in Algorithm~\ref{alg:dmpd}.

\section{Experiments}
\label{sec:experiments}

\begin{table}[t]
    \centering
    \caption{
        Student-only accuracy (\%) on five mathematical reasoning benchmarks.
        ESR uses a fixed-prefix supervision implementation~\citep{ziheng2026less}.
        Subscripts report inference standard errors in 16 independent responses per question.
        Bold / underline mark the best / second-best results per pair.
    }
    \label{tab:main_results}
    \footnotesize
    \setlength{\tabcolsep}{2pt}
    \renewcommand{\arraystretch}{1.03}
    \newcommand{\accse}[2]{\ensuremath{#1_{\scriptscriptstyle{\color{black!55}\pm #2}}}}
    \newcommand{\bestaccse}[2]{\ensuremath{\mathbf{#1}_{\scriptscriptstyle{\color{black!55}\pm #2}}}}
    \newcommand{\secondaccse}[2]{\ensuremath{\underline{#1}_{\scriptscriptstyle{\color{black!55}\pm #2}}}}
    \begin{tabular}{@{}lrrrrrr@{}}
        \toprule
        \textbf{Method}                             & \textbf{AIME24}           & \textbf{AIME25}           & \textbf{AMC23}            & \textbf{AMO}             & \textbf{HMMT}             & \textbf{Average}          \\
        \midrule
        \rowcolor{black!9}
        \multicolumn{7}{@{}c}{\textit{JustRL-DeepSeek-1.5B $\rightarrow$ DeepSeek-R1-Distill-Qwen-1.5B}}                                                                                                                   \\
        \midrule
        Student                                     & \accse{29.58}{2.08}       & \accse{21.88}{1.89}       & \accse{72.34}{1.77}       & \accse{1.44}{0.48}       & \accse{16.10}{1.60}       & \accse{28.27}{0.74}       \\
        +GRPO~\citep{shao2024deepseekmath}          & \accse{31.04}{2.11}       & \accse{24.58}{1.97}       & \accse{71.88}{1.78}       & \accse{1.44}{0.48}       & \accse{13.83}{1.50}       & \accse{28.55}{0.75}       \\
        +OPD~\citep{li2026rethinkingopd}            & \secondaccse{46.25}{2.28} & \bestaccse{33.96}{2.16}   & \secondaccse{87.03}{1.33} & \secondaccse{2.08}{0.57} & \secondaccse{22.92}{1.83} & \secondaccse{38.45}{0.78} \\
        \rowcolor{dmpdnegative!18}
        \textbf{+ESR}~\citep{ziheng2026less} & \accse{37.92}{2.21}       & \accse{28.75}{2.07}       & \accse{78.44}{1.63}       & \accse{1.76}{0.53}       & \accse{16.86}{1.63}       & \accse{32.74}{0.77}       \\
        \rowcolor{dmpddelta!18}
        \textbf{+ATI-OPD}                           & \bestaccse{49.17}{2.28}   & \secondaccse{33.54}{2.15} & \bestaccse{88.59}{1.26}   & \bestaccse{2.72}{0.65}   & \bestaccse{24.24}{1.87}   & \bestaccse{39.65}{0.78}   \\
        \midrule
        \rowcolor{black!9}
        \multicolumn{7}{@{}c}{\textit{JustRL-Nemotron-1.5B $\rightarrow$ OpenMath-Nemotron-1.5B}}                                                                                                                          \\
        \midrule
        Student                                     & \accse{59.17}{2.24}       & \accse{48.75}{2.28}       & \accse{90.78}{1.14}       & \accse{6.09}{0.96}       & \accse{30.49}{2.00}       & \accse{47.06}{0.81}       \\
        +GRPO~\citep{shao2024deepseekmath}          & \accse{59.58}{2.24}       & \accse{48.96}{2.28}       & \accse{90.62}{1.15}       & \accse{5.29}{0.90}       & \accse{30.87}{2.01}       & \accse{47.07}{0.81}       \\
        +OPD~\citep{li2026rethinkingopd}            & \secondaccse{67.08}{2.14} & \secondaccse{59.38}{2.24} & \secondaccse{95.00}{0.86} & \secondaccse{9.78}{1.19} & \accse{32.39}{2.04}       & \secondaccse{52.72}{0.80} \\
        \rowcolor{dmpdnegative!18}
        \textbf{+ESR}~\citep{ziheng2026less} & \accse{63.96}{2.19}       & \accse{54.79}{2.27}       & \accse{93.91}{0.95}       & \accse{6.57}{0.99}       & \secondaccse{32.77}{2.04} & \accse{50.40}{0.80}       \\
        \rowcolor{dmpddelta!18}
        \textbf{+ATI-OPD}                           & \bestaccse{68.75}{2.12}   & \bestaccse{60.42}{2.23}   & \bestaccse{95.63}{0.81}   & \bestaccse{10.90}{1.25}  & \bestaccse{34.66}{2.07}   & \bestaccse{54.07}{0.80}   \\
        \bottomrule
    \end{tabular}
\end{table}

We first examine whether ATI-OPD improves distribution alignment, optimization stability, and response lengths, then evaluate the student's independent reasoning.
Next, we test the roles of local teacher guidance and supervision selection.
Finally, we analyze which interventions improve answers and how teacher participation changes during training.

\subsection{Experimental setup}
\label{sec:setup}

\paragraph{Models and comparison settings.}
We train on DAPO-Math-17K~\citep{yu2025dapo} with a maximum response length of $N_{\max}=23{,}552$ tokens, using two teacher $\rightarrow$ student pairs:
JustRL-DeepSeek-1.5B~\citep{he2025justrl} $\rightarrow$ DeepSeek-R1-Distill-Qwen-1.5B~\citep{deepseekai2025deepseekr1},
and JustRL-Nemotron-1.5B~\citep{he2025justrl} $\rightarrow$ OpenMath-Nemotron-1.5B~\citep{moshkov2025aimo2}.
We refer to these pairs as JustRL-DeepSeek and JustRL-Nemotron, respectively.
Our main configuration permits $M=1$ intervention and applies OPD to retained student and teacher tokens.
Vanilla OPD uses the same training settings as ATI-OPD. See Appendix~\ref{app:hyperparameters} for detailed configurations.

\begin{figure}[!t]
    \centering
    \includegraphics[width=0.9\linewidth]{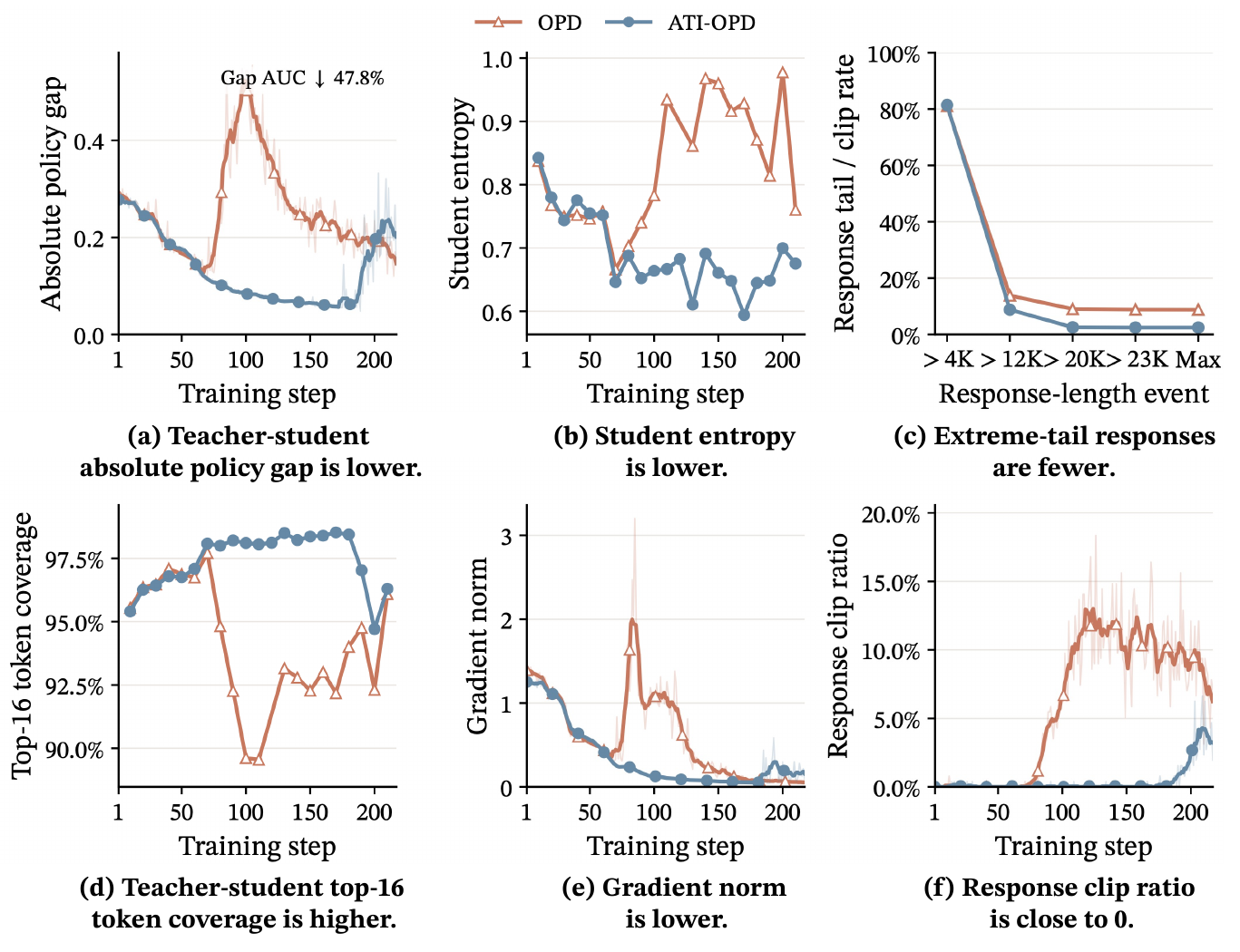}
    \caption{
        Training diagnostics for JustRL-DeepSeek: absolute gap, entropy, response-length tail (last 32 training steps), top-16 coverage, gradient norm, and clipping rate.
    }
    \label{fig:long_horizon_deepseek}
\end{figure}

\vspace{-5pt}
\paragraph{Evaluation and diagnostic metrics.}
We evaluate on AIME 2024~\citep{aime24}, AIME 2025~\citep{aime25}, AMC 2023~\citep{maa2023amc}, AMO-Bench~\citep{liu2026amobench}, and HMMT February 2026~\citep{dekoninck2026matharena}.
We average final-answer correctness over $R=16$ student-only responses per question, then weight the five tasks equally.
Training stability diagnostics include sampled-token absolute gap, the absolute difference between teacher and student log probabilities of each response token, and teacher top-16 token coverage, the fraction of valid tokens among the teacher's 16 most likely choices under the same prefix.
Length and optimization diagnostics include the response-length tail, maximum-length clipping rate, gradient norm, and student entropy.
Detailed definitions for these metrics are in Appendices~\ref{app:stability_metrics} and~\ref{app:curve_aggregation}.

\subsection{Training stability and student-only accuracy}
\label{sec:main_results}
\label{sec:stability}
\label{sec:long_horizon}

\paragraph{Distribution alignment and optimization stability improve together under ATI-OPD.}
We track prediction disagreement and optimization to examine how teacher guidance affects the stability of training.
On JustRL-DeepSeek, the absolute gap initially falls under both methods, then rebounds under vanilla OPD as top-16 coverage drops, as shown in Figure~\ref{fig:long_horizon_deepseek}(a,d).
Entropy increases and gradient spikes occur around the same stage, as shown in Figure~\ref{fig:long_horizon_deepseek}(b,e).
ATI-OPD maintains a lower gap and higher coverage for longer, reducing the raw gap AUC by $47.8\%$, with steadier gradients and lower, less variable entropy.
These improvements support stabilizing training by revising generated prefixes and selecting where to apply supervision.

\FloatBarrier
\Needspace{18\baselineskip}
\begin{wrapfigure}[13]{R}{0.45\textwidth}
    \vspace{-\intextsep}
    \centering
    \captionsetup{skip=0pt}
    \includegraphics[width=0.88\linewidth]{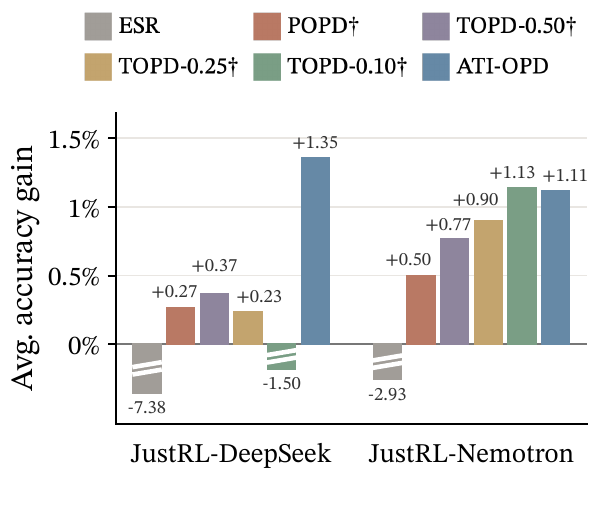}
    \vspace{-8pt}
    \caption{
        Gains over each OPD baseline on three shared tasks (\%).
        $\dagger$: published results.
    }
    \label{fig:cross_paper_compare}
\end{wrapfigure}

\paragraph{ATI-OPD concentrates length reduction in the extreme response tail.}
We compare length thresholds to distinguish suppression of extreme responses from a general loss of long reasoning.
The methods have similar fractions above 4K tokens, but $13.82\%$ of OPD responses exceed 12K over the last 32 steps, compared with $8.78\%$ for ATI-OPD, as shown in Figure~\ref{fig:long_horizon_deepseek}(c).
The difference persists further into the tail, and ATI-OPD reaches the token limit less often, as shown in Figure~\ref{fig:long_horizon_deepseek}(f).
These are complete generated responses, including tokens excluded from supervision.
ATI-OPD thus preserves substantial multi-thousand-token reasoning while reducing the extreme lengths associated with unstable training.

\vspace{-6pt}
\paragraph{Training-time teacher guidance transfers to student-only reasoning.}
Without teacher intervention at evaluation, ATI-OPD reaches $39.65\%$ and $54.07\%$ average accuracy on the two pairs, compared with $38.45\%$ and $52.72\%$ for vanilla OPD, as shown in Table~\ref{tab:main_results}.
The absolute gains are $1.20\%$ and $1.35\%$, with improvements on four of five tasks for JustRL-DeepSeek and all five for JustRL-Nemotron.
ATI-OPD also improves accuracy with a 7B student, as shown in Appendix~\ref{app:larger_student}, demonstrating the scalability of ATI-OPD in terms of model scale.
ESR underperforms ATI-OPD on both pairs, suggesting that its fixed supervision cutoff discards useful learning signals from later reasoning.
On three shared benchmarks, ATI-OPD gains $1.35\%$ and $1.11\%$ over its own OPD baseline, compared with published TOPD/POPD gains of $-1.50\%$ to $+0.37\%$ and $+0.50\%$ to $+1.13\%$, respectively, as shown in Figure~\ref{fig:cross_paper_compare}.
It achieves a larger gain on JustRL-DeepSeek and approaches the best gain on JustRL-Nemotron.
Detailed results are in Appendix~\ref{app:external_comparison}.

\subsection{Mechanisms of intervention and supervision}
\label{sec:state_intervention}
\label{sec:learning_sources}
\label{sec:ablations}

\begin{figure}[!t]
    \centering
    \includegraphics[width=0.9\linewidth]{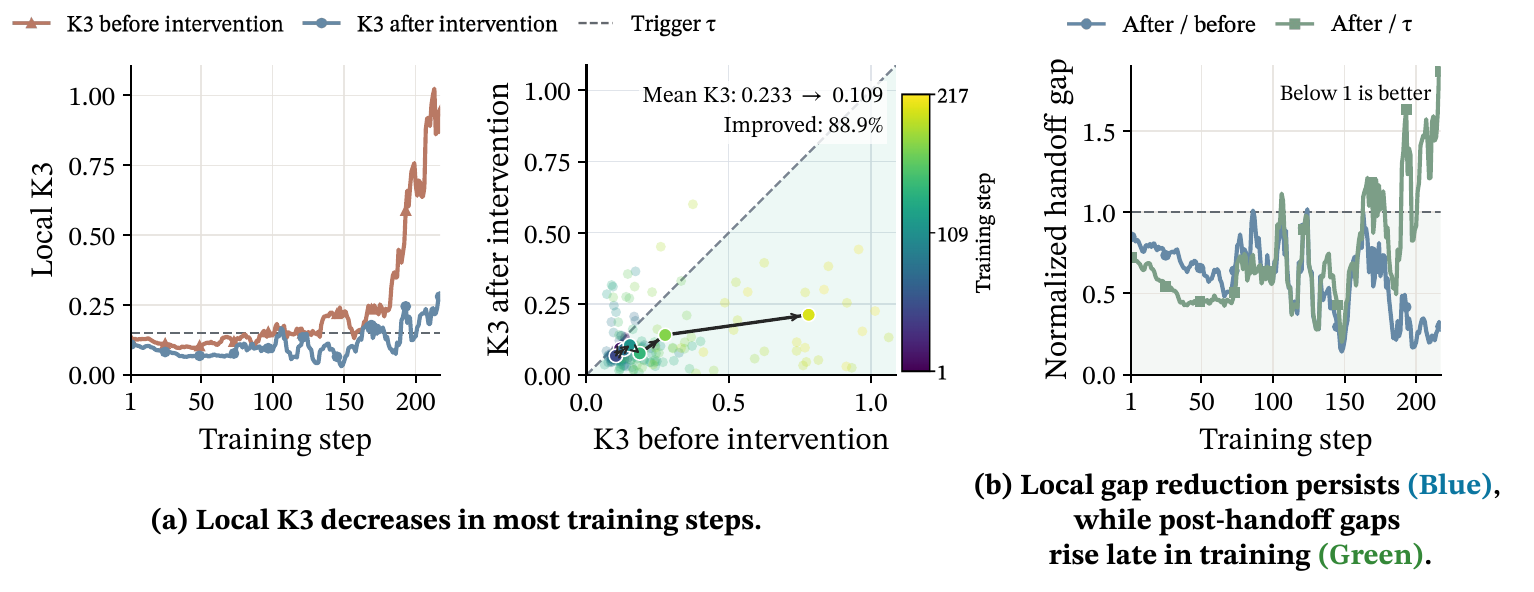}
    \caption{
        Local handoff diagnostics for JustRL-DeepSeek: (a) student K3 after the original and replacement segments, shown as curves and a scatter plot; (b) post-handoff gap ratios.
        Appendix~\ref{app:intervention_metrics} defines the matched student windows.
    }
    \label{fig:repair_gap_dynamics}
    \label{fig:local_gap_scatter}
\end{figure}

\paragraph{Local teacher rewrites improve alignment in the student's own continuation.}
To test whether guidance extends beyond the inserted teacher tokens, we compare student continuations immediately after the original triggering segment and its replacement, matching up to 16 valid tokens per side.
The mean paired decrease in sampled-token K3 is $0.125$ across training steps, with a decrease in $88.9\%$ of training steps.
This improvement occurs in the student's subsequent decisions, consistent with the rewrite providing a better starting point.
The independently measured post-handoff gap nevertheless exceeds $\tau$ late in training, as shown in Figure~\ref{fig:repair_gap_dynamics}, supporting continued monitoring after local guidance.
Appendix~\ref{app:intervention_metrics} gives the paired statistics and the scope of each diagnostic.

\begin{figure}[!t]
    \centering
    \includegraphics[width=0.9\linewidth]{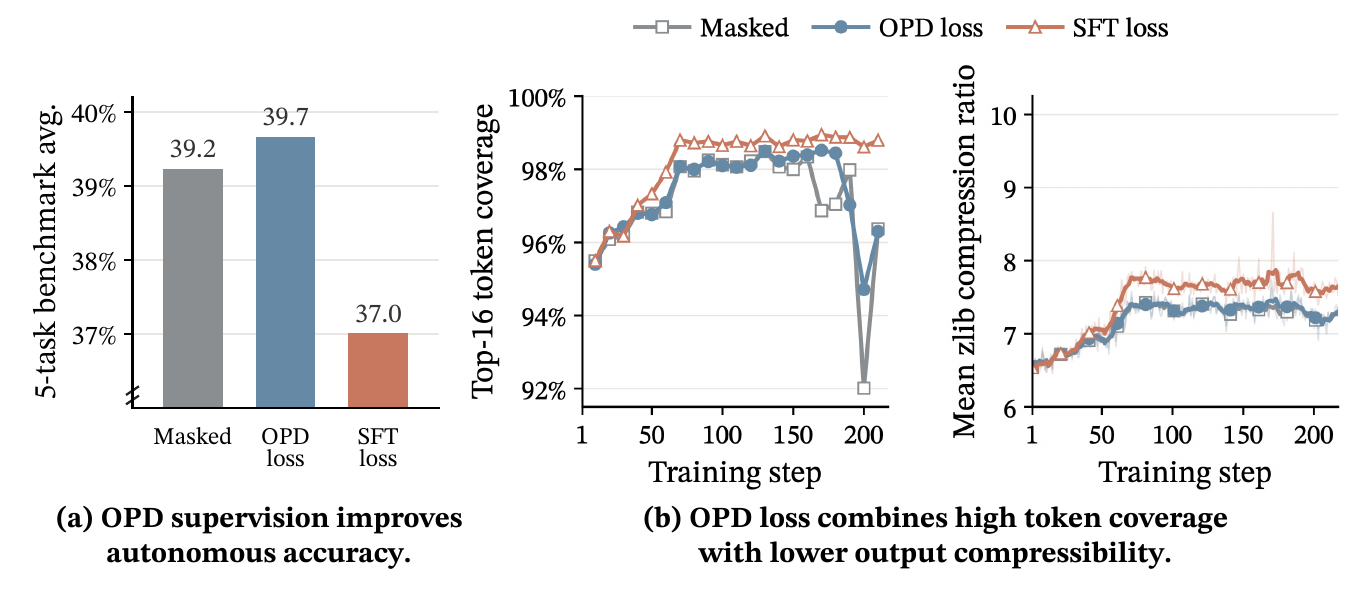}
    \caption{
        Teacher-segment supervision for JustRL-DeepSeek: (a) five-task average accuracy; (b) top-16 coverage and token-zlib compression ratio (higher values indicate greater compressibility).
        Both diagnostics include teacher tokens for OPD, but exclude them for Masked and SFT.
    }
    \label{fig:loss_ablation_diagnostics}
\end{figure}

\paragraph{OPD loss is more effective than SFT for teacher-segment supervision.}
Teacher segments provide both context for continuation and demonstration text, making SFT cross-entropy on the teacher's sampled tokens a natural alternative at these positions.
To examine how supervision at these positions affects training, we compare OPD, Masked, and SFT loss while keeping all inserted tokens as context.
Masked retains teacher context but removes its direct loss, applying OPD only to retained student positions.
Masked achieves $39.22\%$ average accuracy across five tasks, above vanilla OPD's $38.45\%$, as shown in Figure~\ref{fig:loss_ablation_diagnostics}(a).
Retaining the OPD loss on teacher segments increases accuracy to $39.65\%$, compared with $36.99\%$ for SFT cross-entropy.
SFT loss has higher token coverage than OPD loss ($+2.75\%$), but also a higher mean zlib compression ratio ($+0.39$), indicating a greater tendency to repeat, as shown in Figure~\ref{fig:loss_ablation_diagnostics}(b).
These results favor using teacher probabilities across candidate tokens over fitting only the sampled teacher token.

\begin{figure}[!htbp]
    \centering
    \includegraphics[width=0.9\linewidth]{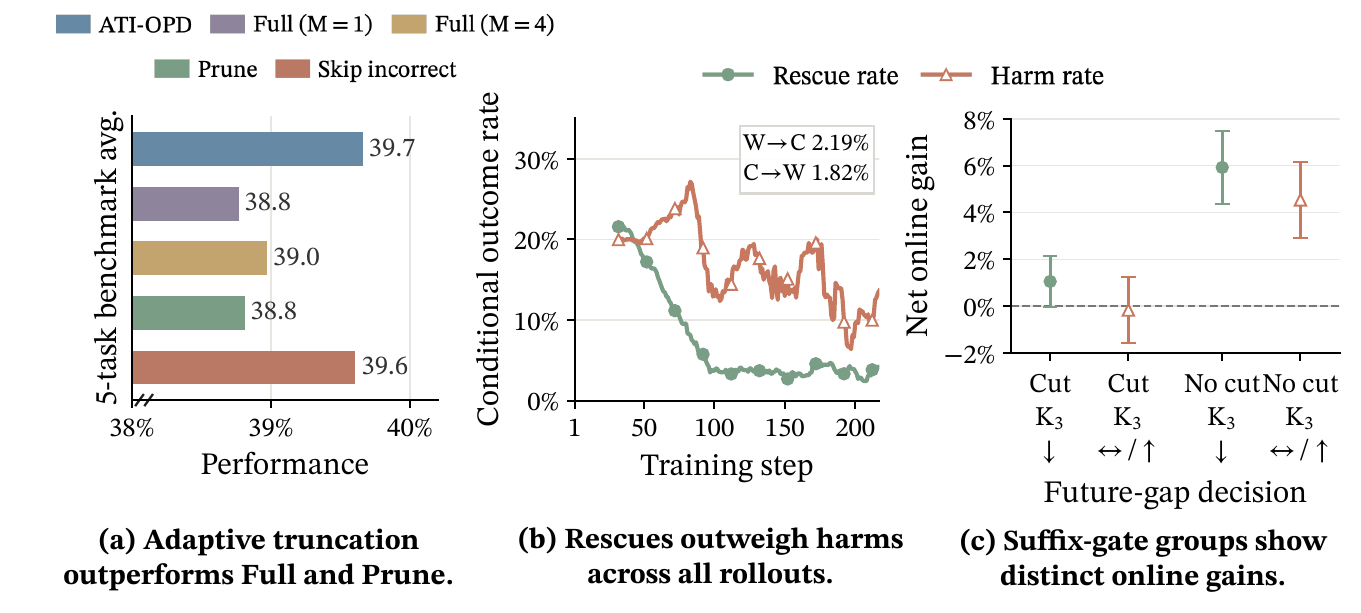}
    \caption{
        JustRL-DeepSeek ablations and outcomes: (a) benchmark accuracy; (b) rescue/harm rates among originally incorrect/correct intervened responses, respectively; inset percentages use all rollouts.
        (c) Online gain (\%) by supervision cutoff (Cut/No cut) and 128-token local K3 change ($\downarrow$: decrease; $\leftrightarrow/\uparrow$: no decrease).
        Error bars are step-cluster bootstrap 95\% confidence intervals; see Appendix~\ref{app:online_outcomes} for details.
    }
    \label{fig:ablation_benchmark}
\end{figure}

\paragraph{Student continuation provides useful supervision when retained selectively.}
We compare ATI-OPD with Full, which supervises the entire continuation, and Prune, which ends supervision at the teacher segment, to test the value of selecting later student decisions.
With $M=1$, ATI-OPD reaches $39.65\%$ five-task accuracy, compared with $38.76\%$ for Full and $38.80\%$ for Prune, as shown in Figure~\ref{fig:ablation_benchmark}(a).
Increasing Full's budget to $M=4$ yields $38.96\%$.
The two extremes both perform worse, indicating that useful student supervision remains after the handoff and that additional interventions do not replace selecting its extent.
Short teacher segments retain the gains: $L_T=128$ and $512$ achieve $39.65\%$ and $39.34\%$, and the tested trigger thresholds span $0.32\%$. See Appendices~\ref{app:teacher_length_sensitivity} and~\ref{app:trigger_threshold_sensitivity} for details.

\subsection{Outcomes and dynamics of intervention}
\label{sec:training_dynamics}

\paragraph{Retaining responses with incorrect final answers still preserves student-only accuracy.}
We compare answer transitions and correctness-based filtering to assess whether final-answer errors justify discarding an entire training trajectory.
Rescues (W$\to$C) and harms (C$\to$W) account for $2.19\%$ and $1.82\%$ of all rollouts, respectively, yielding a net immediate accuracy change, or online gain, of $+0.37\%$, as shown in Figure~\ref{fig:ablation_benchmark}(b).
Skip incorrect retains ATI-OPD's supervision selection but excludes an entire intervened trajectory from training if its final answer is incorrect.
Its $39.60\%$ five-task accuracy is close to ATI-OPD's $39.65\%$, as shown in Figure~\ref{fig:ablation_benchmark}(a).
Final-answer correctness evaluates the whole response, whereas OPD learns from selected token predictions within it.
See Appendix~\ref{app:online_outcomes} for definitions.
A response can therefore retain useful token-level supervision even when its final answer is incorrect.

\paragraph{Persistent suffix divergence separates intervention gains more clearly than immediate local K3 changes.}
To assess whether later student decisions add information beyond the local change, we group responses by supervision cutoff and the 128-token local K3 change, as shown in Figure~\ref{fig:ablation_benchmark}(c).
Without a cutoff, online gain is $+5.92\%$ when local K3 decreases and $+4.54\%$ when it does not; the corresponding values with a cutoff are $+1.05\%$ and $-0.17\%$.
Differences between cutoff groups ($4.7\%$--$4.9\%$) exceed those between local K3 categories ($1.2\%$--$1.4\%$).
Higher gains concentrate on continuations without persistent disagreement, making later divergence a more informative diagnostic of the intervention outcome.

\begin{figure}[!htbp]
    \centering
    \includegraphics[width=0.9\linewidth]{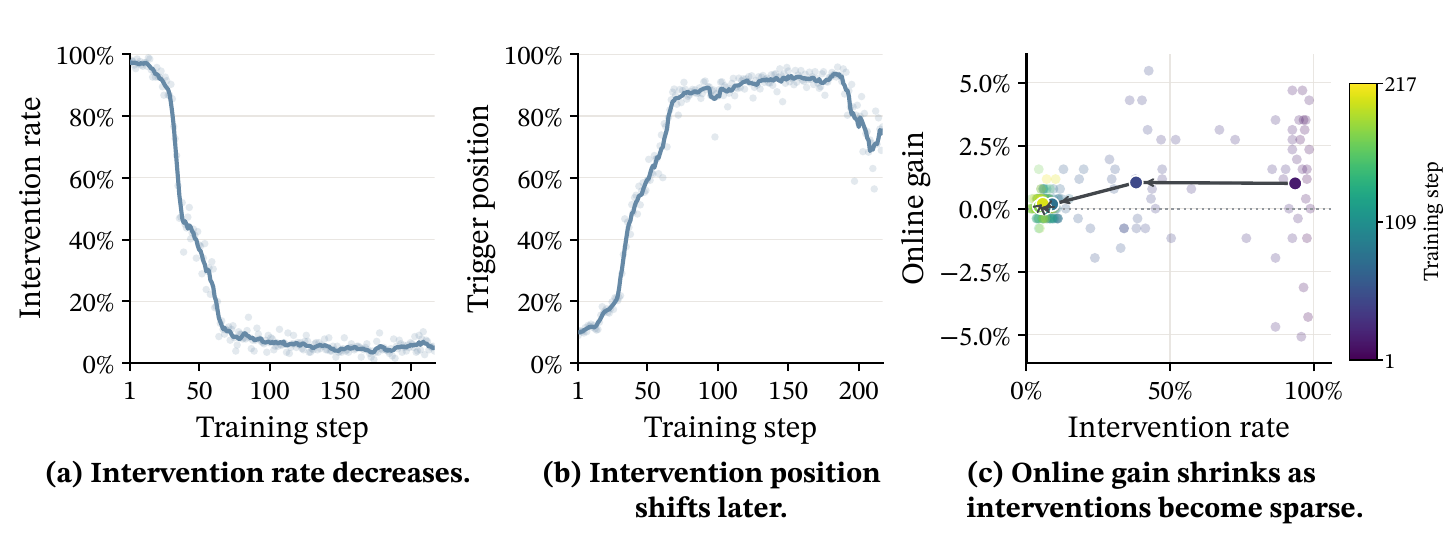}
    \caption{
        Intervention dynamics for JustRL-DeepSeek: (a) intervention rate; (b) first trigger position relative to the original response length; (c) intervention rate versus online gain over all rollouts (\%).
        Colors mark training steps; arrows connect consecutive 32-step phase centers.
    }
    \label{fig:intervention_dynamics}
\end{figure}

\paragraph{Teacher guidance evolves from broad participation to selective intervention on failed trajectories.}
We track frequency, location, and original correctness to examine how the same intervention rule changes its focus during training.
Between the first and last 32 steps, intervention rates fall from $93.57\%$ to $5.51\%$, while original accuracy among intervened responses drops from $49.34\%$ to $9.76\%$, as shown in Figure~\ref{fig:intervention_dynamics}(a).
The first trigger shifts from $15.1\%$ to $79.0\%$ of the original response length, as shown in Figure~\ref{fig:intervention_dynamics}(b).
Guidance thus increasingly targets a small subset of originally incorrect responses at later positions.
Late observations also concentrate at smaller overall online gains: as interventions become sparse, fewer responses can benefit from an immediate answer change, as shown in Figure~\ref{fig:intervention_dynamics}(c).

\FloatBarrier
\section{Limitations}
\label{sec:limitations}

ATI-OPD adds teacher decoding, student regeneration, and scoring costs.
Selecting supervision after full generation does not reduce these decoding costs.
Our main evaluation covers two 1.5B teacher--student pairs in mathematical reasoning, with one training run per setting.
An exploratory 7B-student comparison is limited to four tasks, see Appendix~\ref{app:larger_student} for details, leaving broader model and task coverage for future work.

\section{Conclusion}
\label{sec:conclusion}

We introduce ATI-OPD to improve the stability of long-sequence on-policy distillation.
Adaptive teacher intervention changes the prefix for subsequent student reasoning, and adaptive supervision selection limits learning from renewed disagreement after the handoff.
ATI-OPD reduces distribution mismatch, gradient spikes, and extremely long responses.
Local and supervision ablations support learning selectively from the student's continuation after guidance.
ATI-OPD also improves average student-only accuracy across five mathematical reasoning benchmarks on both model pairs.

\subsection*{AI use statement}

We used AI to give experimental design suggestions, implement the code, write visualization scripts, and interpret experimental results.
AI also helped with manuscript translation, drafting, and editing.
DeepSeek provided additional feedback on exposition and manuscript organization.

We did not use AI to develop theoretical models or conceptual frameworks.
AI use for synthetic dataset generation, dataset cleaning or reformatting, hypothesis formulation or refinement, formulating mathematical claims, providing key proof elements, or writing proofs was not applicable.
We manually reviewed AI-assisted work, including implementation code, visualization scripts, and experimental analyses, and take responsibility for the final paper's content and artifacts.

\subsection*{Ethics statement}

This study trains and evaluates released models on public mathematical reasoning data, with evaluation based on model-generated solutions.
Models, datasets, and outputs are used solely for academic research.
Distillation may transfer teacher errors or biases to the student. Deployment in specific applications requires further assessment of output reliability, safety, and potential misuse.

\subsection*{Reproducibility statement}

Section~\ref{sec:dmpd} describes ATI-OPD's divergence monitoring, adaptive teacher intervention, and supervision selection rules. Appendix~\ref{app:trajectory_implementation} gives the complete trajectory construction procedure for the main configuration in Algorithm~\ref{alg:dmpd}.
Section~\ref{sec:setup} and Appendix~\ref{app:hyperparameters} specify the model and data sources, training and evaluation settings, and ablation variants.
Appendix~\ref{app:metric_definitions} details the optimization objective, evaluation and diagnostic metrics, and aggregation across training steps.

\end{CJK}

\bibliographystyle{./refstyle}
\bibliography{iclr2027_conference}

\appendix
\clearpage
\section{Hyperparameters and model coverage}
\label{app:hyperparameters}

Table~\ref{tab:hyperparameters} summarizes the detailed settings for ATI-OPD.
Vanilla OPD shares all settings except those for teacher intervention and supervision selection.
Here, $\iota_T$ controls OPD on teacher tokens.

\begin{table}[h]
    \centering
    \caption{Main settings for ATI-OPD training, student-only evaluation, and local intervention.}
    \label{tab:hyperparameters}
    \small
    \setlength{\tabcolsep}{6pt}
    \begin{tabular}{@{}ll@{}}
        \toprule
        \textbf{Component}           & \textbf{Setting}                               \\
        \midrule
        Training data                & DAPO-Math-17K                                  \\
        Max prompt length            & 1,024 tokens                                   \\
        Max response length          & $N_{\max}=23{,}552$ tokens                      \\
        Batch size                   & 64                                             \\
        Rollouts per prompt          & 4                                              \\
        Training steps               & 217                                            \\
        Optimizer                    & AdamW, constant learning rate $10^{-6}$        \\
        Weight decay                 & 0.01                                           \\
        Gradient clip                & 1.0                                            \\
        Student rollout              & temperature 1.0, top-$p$ 1.0                   \\
        Evaluation                   & $R=16$, temperature 0.7, top-$p$ 0.95          \\
        OPD candidate set            & Student top-16                                 \\
        Student monitor segment      & $L_S=128$                                      \\
        Teacher intervention segment & $L_T=128$                                      \\
        Trigger                      & Mean clipped divergence score, $\tau=0.15$    \\
        Teacher intervention sampling      & temperature 0.6, top-$p$ 0.95, top-$k$ 20      \\
        Teacher samples per intervention   & 1                                              \\
        Intervention budget                & $M=1$                                          \\
        Future window                & $L_F=64$ tokens, at least 32 valid tokens       \\
        Reference pool               & 2,048 window means from correct original responses \\
        Reference contribution       & At most 8 windows per correct response          \\
        Initial reference threshold  & $\eta_s=\tau$ until 128 windows are available    \\
        Future cutoff                & $\eta_s$: delayed 90th percentile, persistence 2 \\
        Teacher-segment objective    & OPD loss, $\iota_T=1$                          \\
        \bottomrule
    \end{tabular}
\end{table}

\paragraph{Ablation settings.}
\label{app:ablation_variants}
Teacher-segment OPD loss is the main method, applying distribution supervision to both retained student and teacher tokens ($\iota_T=1$).
Teacher-segment masking (Masked) sets $\iota_T=0$, retaining teacher tokens as context but removing their direct loss. The remaining retained student tokens still receive OPD supervision.
The SFT loss variant replaces OPD on teacher tokens with teacher-forcing cross-entropy weighted by $\lambda_{\mathrm{SFT}}=0.1$; Appendix~\ref{app:exact_opd_surrogate} gives the objective.
Full supervises the complete continuation after intervention, with intervention budgets of $M=1$ and $M=4$; Prune restricts supervision to positions through the teacher segment.
Both variants still generate the complete continuation.
The current adaptive supervision implementation requires $M=1$.
Skip incorrect follows ATI-OPD's intervention and supervision selection, then excludes an entire intervened trajectory from training if its final answer is incorrect.

\paragraph{ESR implementation.}
For ESR, we disable teacher intervention and set a fixed 100-token OPD supervision horizon~\citep{ziheng2026less}.
The student still generates responses up to the 23,552-token limit.
A separate mask restricts teacher-derived token rewards and the actor policy loss to the first 100 valid response tokens.

\section{Additional comparisons and intervention analysis}
\label{app:benchmark_trajectories}

\subsection{Exploratory evaluation with a larger student}
\label{app:larger_student}

\paragraph{Exploratory results with a larger student.}
To examine whether the main result extends beyond the 1.5B students, we compare OPD and ATI-OPD with DeepSeek-R1-Distill-Qwen-7B as the student and JustRL-DeepSeek-1.5B as the teacher.
The teacher is unchanged in size.
ATI-OPD is higher on all four available tasks, with a four-task average of $44.72\%$ compared with $43.40\%$ for OPD.

\begin{table}[!htbp]
    \centering
    \caption{
        Exploratory student-only accuracy (\%) with a 7B student.
        Evaluation uses temperature 1.0, top-$p=1.0$, no top-$k$ truncation, with 16 responses per question. Avg. equally weights the four available tasks.
    }
    \label{tab:larger_student}
    \footnotesize
    \setlength{\tabcolsep}{5pt}
    \renewcommand{\arraystretch}{1.03}
    \begin{tabular}{@{}lrrrrr@{}}
        \toprule
        \textbf{Method} & \textbf{AIME24} & \textbf{AIME25} & \textbf{AMC23} & \textbf{AMO-Bench} & \textbf{Avg.} \\
        \midrule
        OPD & 46.46 & 37.71 & 86.41 & 3.04 & 43.40 \\
        ATI-OPD & 48.54 & 39.17 & 87.66 & 3.53 & 44.72 \\
        \bottomrule
    \end{tabular}
\end{table}

\FloatBarrier

\subsection{Contextual comparison with published length control}
\label{app:external_comparison}

We compare ATI-OPD with published length control methods~\citep{zhang2026fullrollouts}.
Table~\ref{tab:cross_paper_gain} reports gains over each study's own vanilla OPD baseline on three shared benchmarks.
The published results average the final ten evaluations, whereas ours use the final training checkpoint.
The original response limit is 15,360 tokens, compared with 23,552 in our experiments.
The TOPD and POPD results show that controlling training trajectory length can improve distillation, with gains varying across tasks and configurations.
ATI-OPD achieves a larger average gain on JustRL-DeepSeek and an average gain close to that of the best length control configuration on JustRL-Nemotron.

\begin{table}[!htbp]
    \centering
    \caption{
        Absolute accuracy gains over each method's own OPD baseline on three shared benchmarks (\%).
        The number following TOPD denotes the truncation length as a fraction of the original maximum response length.
        $\dagger$ denotes gains calculated from the accuracies published in the corresponding work~\citep{zhang2026fullrollouts}.
    }
    \label{tab:cross_paper_gain}
    \footnotesize
    \setlength{\tabcolsep}{4.2pt}
    \renewcommand{\arraystretch}{0.96}
    \begin{tabular}{@{}lrrrr@{}}
        \toprule
        \textbf{Method}    & $\boldsymbol{\Delta}$\textbf{AIME24} & $\boldsymbol{\Delta}$\textbf{AIME25} & $\boldsymbol{\Delta}$\textbf{AMC23} & $\boldsymbol{\Delta}$\textbf{Avg.} \\
        \midrule
        \rowcolor{black!9}
        \multicolumn{5}{@{}l}{\textit{JustRL-DeepSeek-1.5B $\rightarrow$ DeepSeek-R1-Distill-Qwen-1.5B}}                                                                            \\
        POPD$^{\dag}$      & +1.70                                & \underline{+0.50}                    & $-1.40$                             & +0.27                              \\
        TOPD-0.50$^{\dag}$ & +1.00                                & \textbf{+0.80}                       & \underline{$-0.70$}                 & \underline{+0.37}                  \\
        TOPD-0.25$^{\dag}$ & \underline{+2.00}                    & $-0.10$                              & $-1.20$                             & +0.23                              \\
        TOPD-0.10$^{\dag}$ & +1.40                                & $-1.80$                              & $-4.10$                             & $-1.50$                            \\
        \textbf{ATI-OPD}   & \textbf{+2.92}                       & $-0.42$                              & \textbf{+1.56}                      & \textbf{+1.35}                     \\
        \midrule
        \rowcolor{black!9}
        \multicolumn{5}{@{}l}{\textit{JustRL-Nemotron-1.5B $\rightarrow$ OpenMath-Nemotron-1.5B}}                                                                                   \\
        POPD$^{\dag}$      & +1.30                                & +0.40                                & $-0.20$                             & +0.50                              \\
        TOPD-0.50$^{\dag}$ & +1.30                                & +0.70                                & +0.30                               & +0.77                              \\
        TOPD-0.25$^{\dag}$ & +1.40                                & +0.60                                & \textbf{+0.70}                      & +0.90                              \\
        TOPD-0.10$^{\dag}$ & \textbf{+2.00}                       & \textbf{+1.10}                       & +0.30                               & \textbf{+1.13}                     \\
        \textbf{ATI-OPD}   & \underline{+1.67}                    & \underline{+1.04}                    & \underline{+0.63}                   & \underline{+1.11}                  \\
        \bottomrule
    \end{tabular}
\end{table}

\subsection{Teacher segment length sensitivity}
\label{app:teacher_length_sensitivity}

\paragraph{Short teacher segments retain the strongest average performance in the tested range.}
We vary the teacher's segment length to test whether more teacher-generated text is needed after an intervention, keeping the student segment length fixed at $L_S=128$.
The five-task averages for $L_T=128$, $512$, and $1024$ are $39.65\%$, $39.34\%$, and $38.87\%$, respectively (Table~\ref{tab:teacher_length_sensitivity}).
Longer replacements therefore bring no average improvement in these runs.
The 512-token configuration performs best on AIME25, so the effect is task dependent; the average result supports returning control to the student after a short teacher segment.

\begin{table}[!htbp]
    \centering
    \caption{
        Sensitivity to teacher segment length for JustRL-DeepSeek (accuracy, \%), with $L_S=128$.
        The first four benchmarks use the training-step-217 evaluation; HMMT February 2026 uses the corresponding standalone checkpoint evaluation. Avg. equally weights the five tasks.
    }
    \label{tab:teacher_length_sensitivity}
    \footnotesize
    \setlength{\tabcolsep}{3.4pt}
    \renewcommand{\arraystretch}{1.03}
    \begin{tabular}{@{}lrrrrrr@{}}
        \toprule
        $L_T$ & \textbf{AIME24} & \textbf{AIME25} & \textbf{AMC23} & \textbf{AMO-Bench} & \textbf{HMMT Feb26} & \textbf{Avg.} \\
        \midrule
        128  & 49.17 & 33.54 & 88.59 & 2.72 & 24.24 & 39.65 \\
        512  & 48.33 & 36.88 & 87.19 & 2.72 & 21.59 & 39.34 \\
        1024 & 46.46 & 36.46 & 85.94 & 2.56 & 22.92 & 38.87 \\
        \bottomrule
    \end{tabular}
\end{table}

\subsection{Trigger threshold sensitivity}
\label{app:trigger_threshold_sensitivity}

\paragraph{The five-task mean changes little across the three tested trigger thresholds.}
The threshold controls how much local disagreement is tolerated before teacher intervention.
We compare $\tau=0.10$, $0.15$, and $0.30$ to assess whether the main result depends on a narrowly chosen value.
Their five-task averages are $39.94\%$, $39.65\%$, and $39.63\%$, respectively (Table~\ref{tab:trigger_threshold_sensitivity}).
The mean decreases slightly as the threshold increases, with a total difference of $0.32\%$, although individual tasks vary more.
With one training run per setting, this comparison describes the observed sensitivity rather than establishing a statistically optimal threshold.

\begin{table}[!htbp]
    \centering
    \caption{
        Sensitivity to the intervention trigger threshold for JustRL-DeepSeek (accuracy, \%).
        Avg. is the equal-weight mean of five benchmarks.
    }
    \label{tab:trigger_threshold_sensitivity}
    \footnotesize
    \setlength{\tabcolsep}{3.4pt}
    \renewcommand{\arraystretch}{1.03}
    \begin{tabular}{@{}lrrrrrr@{}}
        \toprule
        $\tau$ & \textbf{AIME24} & \textbf{AIME25} & \textbf{AMC23} & \textbf{AMO-Bench} & \textbf{HMMT Feb26} & \textbf{Avg.} \\
        \midrule
        0.10 & 47.71 & 37.50 & 86.72 & 3.37 & 24.43 & 39.94 \\
        0.15 & 49.17 & 33.54 & 88.59 & 2.72 & 24.24 & 39.65 \\
        0.30 & 47.71 & 34.58 & 88.91 & 2.88 & 24.05 & 39.63 \\
        \bottomrule
    \end{tabular}
\end{table}

\FloatBarrier

\section{Implemented objectives and metric definitions}
\label{app:metric_definitions}

Our training implementation follows~\citet{li2026rethinkingopd}.
This section defines the implemented OPD objective, training stability and intervention metrics, and mixed-trajectory construction.

\begingroup
\postdisplaypenalty=1000 
\allowdisplaybreaks[1]

\subsection{Exact Top-K OPD surrogate}
\label{app:exact_opd_surrogate}

To provide distribution supervision over the student's high-probability candidates, the OPD surrogate weights candidate log-probability differences by student probabilities:
\begin{align}
    w_t(v)&=
    \frac{\pi_{\theta_{\mathrm{old}}}(v\mid h_t)}
    {\sum_{u\in\mathcal V_t}\pi_{\theta_{\mathrm{old}}}(u\mid h_t)}, \\
    A_t(v)&=w_t(v)
    \left[\log\pi_T(v\mid h_t)-\log\pi_{\theta_{\mathrm{old}}}(v\mid h_t)\right].
    \label{eq:appendix_topk_advantage}
\end{align}
Here, $u,v$ are candidate tokens, and $\mathcal V_t$ contains candidates with finite student and teacher log-probabilities,
drawn from $\mathcal C_t=\operatorname{TopK}(\pi_{\theta_{\mathrm{old}}}(\cdot\mid h_t),K)$.
$\operatorname{TopK}$ returns the $K$ highest-probability tokens given the prefix $h_t$; we use $K=16$.
$w_t(v)$ is the student probability renormalized over $\mathcal V_t$, and $A_t(v)$ is the corresponding weighted log-probability difference. This is the \texttt{student\_p} weighting used in the implementation.
The teacher scores every student top-$K$ candidate, whether or not it also belongs to the teacher's top-$K$ set.

To compute candidate probability ratios between the current student and the rollout student, we first clip the log-ratio for numerical stability:
\begin{equation}
    \rho_t(v)=\exp\!\left(
    \operatorname{clip}\!\left[
    \log\pi_\theta(v\mid h_t)-\log\pi_{\theta_{\mathrm{old}}}(v\mid h_t),-20,20
    \right]\right),
\end{equation}
where $\rho_t(v)$ is the candidate probability ratio, and $\operatorname{clip}(z,a,b)=\min(\max(z,a),b)$ bounds its input within $[a,b]$. Here, the log-ratio is clipped to $[-20,20]$.

To limit the magnitude of policy updates, we use the probability-ratio clipping of Proximal Policy Optimization (PPO)~\citep{schulman2017proximal} and its dual-clip form~\citep{ye2020mastering}. The loss for each candidate token is
\begin{equation}
    \ell_t(v)=
    \begin{cases}
        \min\!\left\{-c_{\mathrm{clip}}A_t(v),
        \max[-\rho_t(v)A_t(v),-\bar\rho_t(v)A_t(v)]\right\}, & A_t(v)<0,\\
        \max[-\rho_t(v)A_t(v),-\bar\rho_t(v)A_t(v)], & A_t(v)\ge 0,
    \end{cases}
    \label{eq:appendix_dual_clip}
\end{equation}
where $\bar\rho_t(v)=\operatorname{clip}(\rho_t(v),1-\epsilon_\ell,1+\epsilon_h)$ is the clipped probability ratio.
$\epsilon_\ell$ and $\epsilon_h$ control the clipping ranges below and above 1, and $c_{\mathrm{clip}}$ provides an additional clipping coefficient for negative advantages.
The main setting uses $\epsilon_\ell=\epsilon_h=0.2$ and $c_{\mathrm{clip}}=3$; $c_{\mathrm{clip}}$ is distinct from the trajectory cutoff position $c$.

To update only the selected supervision positions, we sum the candidate losses at each position and average over tokens selected for OPD:
\begin{equation}
    \mathcal L_{\mathrm{OPD}}
    =\frac{\sum_t m_t^{\mathrm{OPD}}
    \sum_{v\in\mathcal V_t}\ell_t(v)}
    {\sum_t m_t^{\mathrm{OPD}}}.
    \label{eq:appendix_opd_loss}
\end{equation}
Here, $m_t^{\mathrm{OPD}}$ is the supervision mask defined in Equation~\ref{eq:dmpd_mask}, which determines whether the current position contributes to distillation.

To compare supervision objectives on the teacher segment, the ablations change its mask or add an objective.
OPD sets $\iota_T=1$ for teacher-generated tokens.
Masked sets $\iota_T=0$ while retaining these tokens as context for subsequent student tokens.
SFT also excludes teacher-generated tokens from the OPD, task-reward, entropy, and KL masks, and adds a cross-entropy objective:
\begin{align}
    \mathcal L_{\mathrm{SFT}}
    &=-\frac{1}{\sum_t m_t^T}\sum_t m_t^T\log\pi_\theta(\widetilde y_t\mid h_t), \\
    \mathcal L&=\mathcal L_{\mathrm{OPD}}+\lambda_{\mathrm{SFT}}\mathcal L_{\mathrm{SFT}},
    \label{eq:appendix_teacher_sft}
\end{align}
where $m_t^T\in\{0,1\}$ marks only the inserted teacher tokens, and $h_t=(x,\widetilde y_{<t})$ is the mixed-trajectory prefix.
$\lambda_{\mathrm{SFT}}$ weights the cross-entropy term, and $\mathcal L$ is the combined training loss. The SFT ablation uses $\lambda_{\mathrm{SFT}}=0.1$.

\subsection{Token- and trajectory-level stability metrics}
\label{app:stability_metrics}

\paragraph{Response masks and token sources.}
The logged absolute gap, token coverage, and token-zlib compression ratio use the response mask supplied by the training configuration.
Vanilla OPD contains only student-generated tokens.
In the main ATI-OPD configuration, the mask includes student tokens and inserted teacher tokens in the complete mixed response.
It also includes positions beyond the supervision cutoff: that cutoff changes the distillation loss mask, while the response mask and generated text remain unchanged.
Masked and the teacher-segment SFT variant exclude teacher tokens from this diagnostic mask.
Thus, the aggregate curves describe the tokens selected by each configuration, and their differences should not be attributed entirely to changes in the student's own predictions.
The separate local comparison in Appendix~\ref{app:intervention_metrics} excludes teacher tokens in all cases.

\paragraph{Sampled-token absolute gap.}
To measure teacher--student divergence at the prefixes actually visited, we average the absolute sampled-token log-probability difference over valid positions:
\begin{align}
    d_{it}^{\mathrm{abs}}
    &=\left|\log\pi_T(y_{it}\mid h_{it})
    -\log\pi_{\theta_{\mathrm{old}}}(y_{it}\mid h_{it})\right|, \\
    \operatorname{GapAbs}
    &=\frac{\sum_{i,t}m_{it}^{\mathrm{gap}}d_{it}^{\mathrm{abs}}}
    {\sum_{i,t}m_{it}^{\mathrm{gap}}}.
    \label{eq:metric_gap_abs}
\end{align}
Here, $i$ and $t$ index the response and token position, respectively, $y_{it}$ is the actual sampled token, and $h_{it}=(x_i,y_{i,<t})$ consists of prompt $x_i$ and the preceding response tokens.
$d_{it}^{\mathrm{abs}}$ is the absolute difference at one position, and $\operatorname{GapAbs}$ is the aggregate metric.
The validity mask $m_{it}^{\mathrm{gap}}\in\{0,1\}$ is 1 only when the response mask is valid and both log-probabilities are finite, no greater than $10^{-6}$, and at most $10^4$ in absolute value.

\paragraph{Top-16 token coverage.}
To measure how often actual response tokens belong to the teacher's high-probability candidates, we compute teacher top-16 token coverage under the same question and preceding text:
\begin{equation}
    \operatorname{Cov}_{16}
    =\frac{\sum_{i,t}m_{it}^{16}
    \mathbf 1[y_{it}\in\mathcal T_{it}^{16}]}
    {\sum_{i,t}m_{it}^{16}}.
    \label{eq:metric_top16_coverage}
\end{equation}
Here, $\mathcal T_{it}^{16}$ contains the teacher's 16 highest-probability tokens under prefix $h_{it}$, and $\mathbf 1[\cdot]$ is the indicator function.
$m_{it}^{16}$ marks valid positions, requiring a valid response token and at least one finite teacher top-16 log-probability at that position.

\paragraph{Student entropy.}
We use student entropy to track uncertainty in the sampling distribution. To reduce communication in the distributed cluster, we approximate entropy using the top-16 probabilities and a residual bucket containing the remaining probability mass:
\begin{equation}
    \widehat H_{it}^{16}
    =-\sum_{v\in\mathcal S_{it}^{16}}p_{itv}\log p_{itv}
     -r_{it}\log r_{it}.
    \label{eq:metric_entropy}
\end{equation}
Here, $\mathcal S_{it}^{16}$ contains the student's 16 highest-probability tokens under prefix $h_{it}$, and $p_{itv}=\pi_{\theta_{\mathrm{old}}}(v\mid h_{it})$ is the probability of candidate token $v$.
The remaining probability mass is $r_{it}=\operatorname{clip}(1-\sum_{v\in\mathcal S_{it}^{16}}p_{itv},0,1)$, with $0\log0:=0$.

\paragraph{Response tail and clipping.}
To characterize the tail of the response-length distribution, we measure the fraction of responses exceeding each length threshold and the clipping rate at the maximum token budget:
\begin{align}
    S(q)&=\frac{1}{B}\sum_{i=1}^{B}\mathbf 1[N_i>q], \\
    R_{\mathrm{clip}}&=\frac{1}{B}\sum_{i=1}^{B}\mathbf 1[N_i=N_{\max}].
    \label{eq:metric_response_tail}
\end{align}
Here, $B$ is the number of responses in a batch, $N_i$ is the valid length of response $i$, $q$ is the length threshold, and $N_{\max}$ is the maximum token budget.
For ATI-OPD, these are the complete mixed responses after the intervention, including positions beyond the supervision cutoff.
Their lengths exclude discarded text from the initial full rollout and therefore do not measure the total number of tokens generated during training.
$S(q)$ is the survival rate beyond length $q$, and $R_{\mathrm{clip}}$ is the fraction of responses that reach $N_{\max}$.
The ``$N>12$K ratio'' in the main text uses $q=12000$.
The tail plot reports the arithmetic mean of the step-level rates over the final 32 training steps.
Its Max entry uses $R_{\mathrm{clip}}$, whereas other length thresholds use $S(q)$; the former includes responses equal to the length limit.

\paragraph{Token-zlib compression ratio.}
To help diagnose repetition in responses, the supervision-objective ablation uses the logged mean token-zlib compression ratio over the last 4,096 valid tokens of each response, after applying the response mask.
Responses with fewer than 4,096 valid tokens have no compression-ratio observation and are excluded from this mean.
A higher value indicates that the token sequence is more compressible.
The loss-ablation differences in coverage and compressibility are means over the final 32 training steps, using the available observations for each metric.
Coverage is logged less frequently, so its late-stage mean uses three observations in these runs; the compression statistic uses step-level observations.

\paragraph{Gradient norm.}
To monitor gradient spikes during optimization, we record the global $\ell_2$ norm before gradient clipping:
\begin{equation}
    G_s=\left(\sum_{p}\|\nabla_{\theta_p}\mathcal L_s\|_2^2\right)^{1/2}.
    \label{eq:metric_grad_norm}
\end{equation}
Here, $G_s$ is the gradient norm at training step $s$, $p$ indexes parameter tensors, $\theta_p$ denotes the corresponding parameters, and $\mathcal L_s$ is the training loss at that step.

\paragraph{Student-only accuracy.}
To evaluate the trained student's independent reasoning ability, we measure benchmark accuracy without teacher intervention and average across tasks:
\begin{align}
    \operatorname{Acc}_b
    &=\frac{1}{R|\mathcal Q_b|}
    \sum_{q\in\mathcal Q_b}\sum_{r=1}^{R}
    \mathbf 1[\operatorname{judge}(y_{qr})=1], \\
    \operatorname{Avg}&=\frac{1}{|\mathcal B|}\sum_{b\in\mathcal B}\operatorname{Acc}_b.
    \label{eq:metric_accuracy}
\end{align}
Here, $\mathcal B$ is the set of evaluation benchmarks, and $\mathcal Q_b$ is the question set for benchmark $b$.
We sample $R=16$ independent responses per question. $y_{qr}$ is the $r$-th response to question $q$, and $\operatorname{judge}$ returns a correctness label based on the final answer (1 for correct and 0 for incorrect).
$\operatorname{Acc}_b$ is the accuracy on benchmark $b$, and $\operatorname{Avg}$ is the mean accuracy across benchmarks.
The main evaluation set contains AIME24, AIME25, AMC23, AMO-Bench, and HMMT February 2026.
For the small subscripts in Table~\ref{tab:main_results}, we use the conditional Bernoulli plug-in standard error
\begin{align}
    \widehat{\operatorname{SE}}_b
    &=\sqrt{\frac{\operatorname{Acc}_b(1-\operatorname{Acc}_b)}{R|\mathcal Q_b|}}, \\
    \widehat{\operatorname{SE}}_{\mathrm{Avg}}
    &=\frac{1}{|\mathcal B|}
    \left(\sum_{b\in\mathcal B}\widehat{\operatorname{SE}}_b^2\right)^{1/2}.
    \label{eq:metric_accuracy_se}
\end{align}
Both quantities are multiplied by 100 when reported in percentage points.
This calculation treats the sampled responses as independent conditional on a fixed checkpoint and question set; it does not estimate variability across training runs or question sets.
The archived benchmark summaries do not contain the per-question outcomes needed for a paired significance analysis, so we report no benchmark significance tests.

\subsection{Across-step aggregation and plotting conventions}
\label{app:curve_aggregation}

To compare overall divergence during training, we compute the area under the curve (AUC), normalized by the training-step interval, and report the reduction of ATI-OPD relative to OPD:
\begin{align}
    \operatorname{AUC}(z)
    &=\frac{1}{s_J-s_1}\sum_{j=1}^{J-1}
    \frac{z_j+z_{j+1}}{2}(s_{j+1}-s_j), \\
    \operatorname{Red}_{\mathrm{AUC}}
    &=100\frac{\operatorname{AUC}_{\mathrm{OPD}}-
    \operatorname{AUC}_{\mathrm{ATI\text{-}OPD}}}{\operatorname{AUC}_{\mathrm{OPD}}}.
    \label{eq:metric_auc_reduction}
\end{align}
Here, the $J$ observations correspond to training steps $s_1<\cdots<s_J$, and $z_j$ is the raw metric value at observation $j$.
$\operatorname{AUC}_{\mathrm{OPD}}$ and $\operatorname{AUC}_{\mathrm{ATI\text{-}OPD}}$ are the AUC values for the two methods, and $\operatorname{Red}_{\mathrm{AUC}}$ is the relative reduction (\%).
The absolute-gap AUC reduction integrates the raw step-level values from Equation~\ref{eq:metric_gap_abs}.
We integrate over the actual recorded training steps without smoothing or filling missing observations; the trapezoidal rule connects adjacent available points.
For the main comparison, the integration interval is steps 1--217.

\subsection{Intervention location, local effect, and supervised suffix length}
\label{app:intervention_metrics}

To characterize teacher participation and intervention locations, we measure the batch intervention rate, the mean number of interventions per response, and the first trigger position relative to the original response length:
\begin{align}
    r&=\frac{1}{B}\sum_{i=1}^{B}\mathbf 1[K_i>0], \\
    \overline K&=\frac{1}{B}\sum_{i=1}^{B}K_i, \\
    P_{\mathrm{first}}
    &=\frac{1}{|\mathcal I|}\sum_{i\in\mathcal I}\frac{s_i^{(1)}}{\max(N_i^{\mathrm{raw}},1)},
    \label{eq:metric_intervention_rate_position}
\end{align}
where $K_i$ counts completed replacements in response $i$, and $r$ and $\overline K$ are the batch intervention rate and mean intervention count, respectively.
$s_i^{(1)}$ is the zero-based token offset of the first intervention, $N_i^{\mathrm{raw}}$ is the length of the complete original student response, and $\mathcal I=\{i:K_i>0\}$ indexes responses that receive an intervention.
$P_{\mathrm{first}}$ is averaged only over this set and is undefined when no valid intervened response is available.
An intervention is counted when a replacement is completed, regardless of whether the final answer changes from incorrect to correct.

To measure local divergence changes after a teacher replacement, we pair the original student continuation immediately after the triggering segment with the new student continuation immediately after the teacher segment.
We compare means of the clipped divergence score $\bar g_t$ from Equation~\ref{eq:clipped_gap} over up to 16 valid student tokens at the start of each region:
\begin{align}
    G_{\mathrm{before}}&=\frac{1}{n}\sum_{j=1}^{n}b_j, \\
    G_{\mathrm{after}}&=\frac{1}{n}\sum_{j=1}^{n}a_j, \\
    \Delta G&=G_{\mathrm{before}}-G_{\mathrm{after}}.
    \label{eq:metric_local_gap}
\end{align}
Here, $(b_1,\ldots,b_{n_b})$ is the sequence of finite divergence scores following the original triggering segment,
$(a_1,\ldots,a_{n_a})$ is the sequence of finite scores in the student continuation after the teacher segment, and $n=\min(16,n_b,n_a)$.
$G_{\mathrm{before}}$ and $G_{\mathrm{after}}$ are the local means before and after intervention, respectively, and $\Delta G$ is their difference.
Teacher-generated tokens are excluded from both windows.
The two windows follow different versions of the preceding segment; this is a paired descriptive comparison of different continuations.
An intervention contributes only when $n>0$.

For each training step $s$, the logged paired decrease $\Delta G_s$ is the mean of $\Delta G$ over valid intervention pairs.
We report its mean over valid steps and the fraction of steps with a positive decrease:
\begin{equation}
    \frac{1}{|\mathcal S_{\mathrm{valid}}|}
    \sum_{s\in\mathcal S_{\mathrm{valid}}}
    \mathbf 1[\Delta G_s>0],
    \label{eq:metric_improved_steps}
\end{equation}
where $\mathcal S_{\mathrm{valid}}$ is the set of training steps with valid paired statistics.
The mean paired decrease is $0.125$ over 217 steps; the $88.9\%$ decrease rate corresponds to 193 of those steps, not to a fraction of individual interventions.
The paired-decrease curve uses a centered 7-step mean for display; the numerical summary and decrease rate use unsmoothed step values.

We separately track the post-handoff score over up to 16 valid student tokens, without requiring a valid original comparison window.
For comparison with the trigger threshold, we average this score within each step, smooth the step means with a centered 7-step mean, and divide by $\tau$.
This diagnostic can use a different set of interventions and different window lengths from the paired statistic.
It therefore cannot be combined with $\Delta G_s$ to reconstruct a paired original mean or an after-to-before ratio.
The run contains 13,186 intervention boundaries, with 12,949 valid pairs and 13,017 valid independent post-handoff windows.
Of the independent windows, 68 have no original comparison window; eight paired comparisons also use fewer tokens than their independent post-handoff windows.
Paired before/after means and their ratios must therefore be computed from the same matched windows.

To quantify how much of the student suffix receives supervision under Equation~\ref{eq:cutoff_condition}, we compute the supervised suffix length, its fraction of the generated suffix, and an indicator for no supervision cutoff:
\begin{align}
    H_i&=c_i-e_i, \\
    F_i&=
    \begin{cases}
        (c_i-e_i)/(N_i^{\mathrm{gen}}-e_i), & N_i^{\mathrm{gen}}>e_i,\\
        1, & N_i^{\mathrm{gen}}=e_i,
    \end{cases} \\
    U_i&=\mathbf 1[c_i=N_i^{\mathrm{gen}}].
    \label{eq:metric_retained_horizon}
\end{align}
Here, $e_i$ is the final position of the teacher segment, $N_i^{\mathrm{gen}}$ is the complete generated response length, and $c_i$ is the final supervised position.
$H_i$ is the supervised suffix length, $F_i$ is its fraction of the generated suffix, and $U_i$ is the no-cut indicator.
When a cutoff is selected, both detection windows and all later tokens remain in the generated response but are excluded from distillation.
We average $F_i$ and $U_i$ separately over intervened responses.
The ratio $F_i$ describes supervision coverage, not a saving in generation cost.

\subsection{Conditional outcomes and online gain}
\label{app:online_outcomes}

To distinguish the effects of teacher intervention on originally incorrect and correct responses, we score final answers before and after intervention within the same aggregation window and compute the rescue and harm rates:
\begin{align}
    R_{\mathrm{rescue}}&=\frac{n_{WC}}{N_{I,W}}, \\
    R_{\mathrm{harm}}&=\frac{n_{CW}}{N_{I,C}}.
    \label{eq:metric_rescue_harm}
\end{align}
Here, W and C denote incorrect and correct final answers, respectively, and $n_{WC}$ and $n_{CW}$ count transitions from incorrect to correct and from correct to incorrect.
$N_{I,W}$ and $N_{I,C}$ count intervened responses whose original answers are incorrect and correct, respectively.
These are the respective denominators of $R_{\mathrm{rescue}}$ and $R_{\mathrm{harm}}$; the original accuracy of the intervened subset is $N_{I,C}/(N_{I,W}+N_{I,C})$.
The curves aggregate counts over a trailing 32-step window before computing the ratios. A ratio is undefined when its denominator is zero.

To measure the net accuracy change from rescue and harm, we compute online gain for both the intervened subset and all rollouts:
\begin{align}
    \Delta A_I&=100\frac{n_{WC}-n_{CW}}{N_I}, \\
    \Delta A_{\mathrm{all}}&=100\frac{n_{WC}-n_{CW}}{B}.
    \label{eq:online_decomposition}
\end{align}
Here, $N_I=N_{I,W}+N_{I,C}$ is the number of intervened responses, and $B$ is the total number of rollouts in the current aggregation window.
$\Delta A_I$ and $\Delta A_{\mathrm{all}}$ are the respective gains, both expressed in \%.
Responses without intervention remain unchanged and contribute zero to the net change.
Thus, $\Delta A_{\mathrm{all}}=(N_I/B)\Delta A_I$, where $N_I/B$ is the intervention rate.

To show how intervention gains change during training, the Online gain scatter plot in Figure~\ref{fig:intervention_dynamics} uses $\Delta A_{\mathrm{all}}$.
Each point represents a training step, colored by step; arrows connect the centers of consecutive 32-step phases.
Online metrics assess changes in the current training responses, while student-only accuracy after training is evaluated separately by Equation~\ref{eq:metric_accuracy}.

For the four groups in Figure~\ref{fig:ablation_benchmark}(c), each response is grouped by whether the continuation's supervision was cut short and whether its 128-token local divergence mean decreased.
Correctness is evaluated on the complete generated response, including any tokens excluded from distillation.
This local grouping statistic differs from the paired first-16-token diagnostic in Equation~\ref{eq:metric_local_gap}.
Each group's online gain uses that group's response count as its denominator.
The groups contain 12,949 intervened responses with complete grouping information; 237 interventions lack a valid local comparison and are excluded.
Error bars are 95\% percentile intervals from 10,000 bootstrap resamples of training steps, retaining all responses within each sampled step (random seed 20260921).
These intervals describe variation in the recorded training outcomes, not variability across independent training runs or uncertainty in the final benchmark scores.

\endgroup

\subsection{Trajectory construction algorithm and implementation details}
\label{app:trajectory_implementation}

To construct a mixed trajectory for distillation, Algorithm~\ref{alg:dmpd} combines divergence monitoring, teacher intervention, and supervision selection into the complete procedure for the main setting $M=1$.
Its inputs follow the definitions in Section~\ref{sec:dmpd}.

\begin{algorithm}[htbp]
    \caption{Adaptive teacher intervention and supervision selection in ATI-OPD ($M=1$)}
    \label{alg:dmpd}
    \small
    \begin{algorithmic}[1]
        \Require prompt $x$; student $\pi_{\theta_{\mathrm{old}}}$; teacher $\pi_T$
        \Require lengths $L_S,L_T,L_F$; thresholds $\tau,\eta_s$; budget $N_{\max}$
        \Ensure mixed response $\widetilde y$, token owners $o$, final supervised position $c$
        \State sample a raw response $y^{(0)}\sim\pi_{\theta_{\mathrm{old}}}(\cdot\mid x)$ under $N_{\max}$
        \State score eligible segments; $\mathcal K\gets\{k:G(\mathrm{seg}_k^{\mathrm{stu}})>\tau\}$
        \If{$\mathcal K=\emptyset$}
            \State \Return $y^{(0)},S^{|y^{(0)}|},|y^{(0)}|$
        \EndIf
        \State $k^\star\gets\min\mathcal K$; $p\gets p_{<k^\star}$; discard the original suffix
        \State sample $z\sim\pi_T(\cdot\mid x,p)$ with at most $\min(L_T,N_{\max}-|p|)$ tokens
        \State $u\gets\emptyset$
        \If{$p\oplus z$ has not ended and $|p\oplus z|<N_{\max}$}
            \State sample a complete suffix $u\sim\pi_{\theta_{\mathrm{old}}}(\cdot\mid x,p,z)$ under the remaining budget
        \EndIf
        \State $\widetilde y\gets p\oplus z\oplus u$; $o\gets S^{|p|}\oplus T^{|z|}\oplus S^{|u|}$; $c\gets|\widetilde y|$
        \State score $\widetilde y$; split $u$ into windows of at most $L_F$ tokens
        \For{successive windows $j=1,2,\ldots$ in $u$}
            \State compute $Q_j$ if window $j$ is valid; let $b_j$ be its first response position
            \If{$j\ge2$, both adjacent windows are valid, and $Q_{j-1}>\eta_s,\ Q_j>\eta_s$}
                \State $c\gets b_{j-1}-1$
                \State \textbf{break} \Comment{stop the cutoff search; keep the generated response}
            \EndIf
        \EndFor
        \State \Return $\widetilde y,o,c$
    \end{algorithmic}
\end{algorithm}

Here, $S^n$ and $T^n$ denote $n$ identical token-owner markers, and $\oplus$ denotes sequence concatenation.
The algorithm returns the complete mixed response $\widetilde y$, token-owner sequence $o=(o_t)$, and final supervised position $c$.
To apply OPD only at selected positions, we use the following mask over the complete response:
\begingroup
\postdisplaypenalty=10000
\begin{equation}
    m_t^{\mathrm{OPD}}
    =m_t^{\mathrm{base}}\mathbf 1[t\le c]
    \left(\mathbf 1[o_t=S]+\iota_T\mathbf 1[o_t=T]\right).
    \label{eq:dmpd_mask}
\end{equation}
\endgroup
Here, $m_t^{\mathrm{OPD}}$ marks positions used for distillation, and $m_t^{\mathrm{base}}\in\{0,1\}$ marks valid response tokens.
$o_t\in\{S,T\}$ records whether a token was generated by the student or teacher.
$\iota_T\in\{0,1\}$ controls whether teacher tokens receive OPD supervision.

\paragraph{Valid segments and windows.}
To make divergence statistics reflect the student's generation state, monitoring uses only student response tokens with valid log-probabilities, excluding padding and teacher tokens.
For numerical stability, we clip the log-probability difference and the resulting K3 score:
\begin{align}
    \bar\delta_t&=\operatorname{clip}(\delta_t,-20,20), \\
    \bar g_t&=\operatorname{clip}\!\left(\exp(\bar\delta_t)-\bar\delta_t-1,-10,10\right).
    \label{eq:clipped_gap}
\end{align}
Here, $\bar\delta_t$ is the bounded log-probability difference and $\bar g_t$ is the clipped sampled-token K3 score.
Segment triggers, continuation windows, and local intervention diagnostics all use this clipped score; the absolute-gap diagnostic is defined separately in Equation~\ref{eq:metric_gap_abs}.
Both the sums and denominators of segment and window means count only valid tokens, while starting positions retain their original response offsets.
Suffix windows must meet the minimum valid-token count in Table~\ref{tab:hyperparameters}; both adjacent windows in Equation~\ref{eq:cutoff_condition} must be valid and exceed the threshold.
A single threshold crossing does not trigger the cutoff. The reference pool stores valid window means from original student responses with correct final answers, retains the latest values up to the capacity in Table~\ref{tab:hyperparameters}, and sets the threshold to the delayed 90th percentile.
Each correct response contributes at most eight window means, sampled at evenly spaced indices when more are available.
The threshold is $\tau$ until the pool contains 128 windows.
Updates from a completed rollout batch affect only the following batch, so the current response cannot change its own reference threshold.

\paragraph{Budgets and termination.}
To satisfy both the local intervention limit and the total response budget, the teacher generates at most $\min(L_T,N_{\max}-|p_{<k}|)$ tokens per intervention.
If the teacher segment ends the response, the student suffix is $u^{\mathrm{stu}}=\emptyset$; otherwise, the student continues until the response ends or the total budget is exhausted.
All valid generated positions remain eligible for supervision if no cutoff is selected, and the original student rollout is used if no intervention occurs.
The adaptive supervision implementation uses one intervention.
In Full with multiple interventions, the next trigger is sought only after the latest teacher segment, leaving earlier teacher segments intact.

\paragraph{Supervision masking and compute cost.}
The implementation applies the cutoff only to the distillation loss mask, preserving the complete response and its token-owner sequence.
Equation~\ref{eq:cutoff_condition} uses one-based token positions to set $c=b_{j^\star}-1$; the implementation admits zero-based positions strictly below $c$.
Both windows that exceed the threshold and all subsequent positions are excluded from the loss, as in Equation~\ref{eq:dmpd_mask}.
The complete new continuation has already incurred generation and scoring costs.
We therefore do not interpret the supervision cutoff or the excluded-token fraction as decoding savings.

\end{document}